\documentclass[10pt,twocolumn,letterpaper]{article}

\usepackage{iccv}
\usepackage{times}
\usepackage{epsfig}
\usepackage{graphicx}
\usepackage{amsmath}
\usepackage{amssymb}
\usepackage{bm}
\usepackage{booktabs}
\usepackage{colortbl}
\usepackage{color}
\usepackage{xcolor}
\usepackage{array}
\usepackage{arydshln}

\usepackage[breaklinks=true,bookmarks=false]{hyperref}

\iccvfinalcopy 

\def\iccvPaperID{5736} 

\ificcvfinal\fi

\begin{document}

\title{Space Engage: Collaborative Space Supervision for Contrastive-based Semi-Supervised Semantic Segmentation}

\author{Changqi Wang\textsuperscript{\rm 1}\thanks{equal contribution},
        Haoyu Xie\textsuperscript{\rm 1}$^\ast$,
        Yuhui Yuan\textsuperscript{\rm 3},
        Chong Fu\textsuperscript{\rm 1,4}, 
        Xiangyu Yue\textsuperscript{\rm 2} \and
        \textsuperscript{\rm 1} School of Computer Science and Engineering, Northeastern University, Shenyang, China \\
        \textsuperscript{\rm 2} The Chinese University of Hong Kong\quad
        \textsuperscript{\rm 3} Microsoft Research Asia\\
        \textsuperscript{\rm 4} Key Laboratory of Intelligent Computing in Medical Image, Ministry of Education, NEU, China
        \and
        {\tt\small 2101668@stu.neu.edu.cn, xiehaoyu@stumail.neu.edu.cn, yuyua@microsoft.com} \\
        {\tt\small fuchong@mail.neu.edu.cn, xyyue@ie.cuhk.edu.hk}
}

\maketitle
\ificcvfinal\thispagestyle{empty}\fi

\begin{abstract}
   Semi-Supervised Semantic Segmentation (S4) aims to train a segmentation model with limited labeled images and a substantial volume of unlabeled images.
   To improve the robustness of representations, powerful methods introduce a pixel-wise contrastive learning approach in latent space (\textit{i.e.}, representation space) that aggregates the representations to their prototypes in a fully supervised manner.
   However, previous contrastive-based S4 methods merely rely on the supervision from the model's output (logits) in logit space during unlabeled training.
   In contrast, we utilize the outputs in both logit space and representation space to obtain supervision in a collaborative way.
   The supervision from two spaces plays two roles: 1) reduces the risk of over-fitting to incorrect semantic information in logits with the help of representations; 2) enhances the knowledge exchange between the two spaces.
   Furthermore, unlike previous approaches, we use the similarity between representations and prototypes as a new indicator to tilt training those under-performing representations and achieve a more efficient contrastive learning process. 
   Results on two public benchmarks demonstrate the competitive performance of our method compared with state-of-the-art methods.
\end{abstract}

\section{Introduction}
Semantic segmentation is a fundamental task in computer vision, aiming to classify each pixel in an image.
Significant progress \cite{FCN,deeplabv3+} has been made in training on high-quality labeled images using segmentation models composed of an encoder and a segmentation head.
However, annotating images is expensive and time-consuming.
Semi-supervised Semantic Segmentation (S4) alleviates the thirst for annotation by leveraging unlabeled images to train segmentation models.
\begin{figure}[t] 
  \centering
  \includegraphics[width=1.0\linewidth]{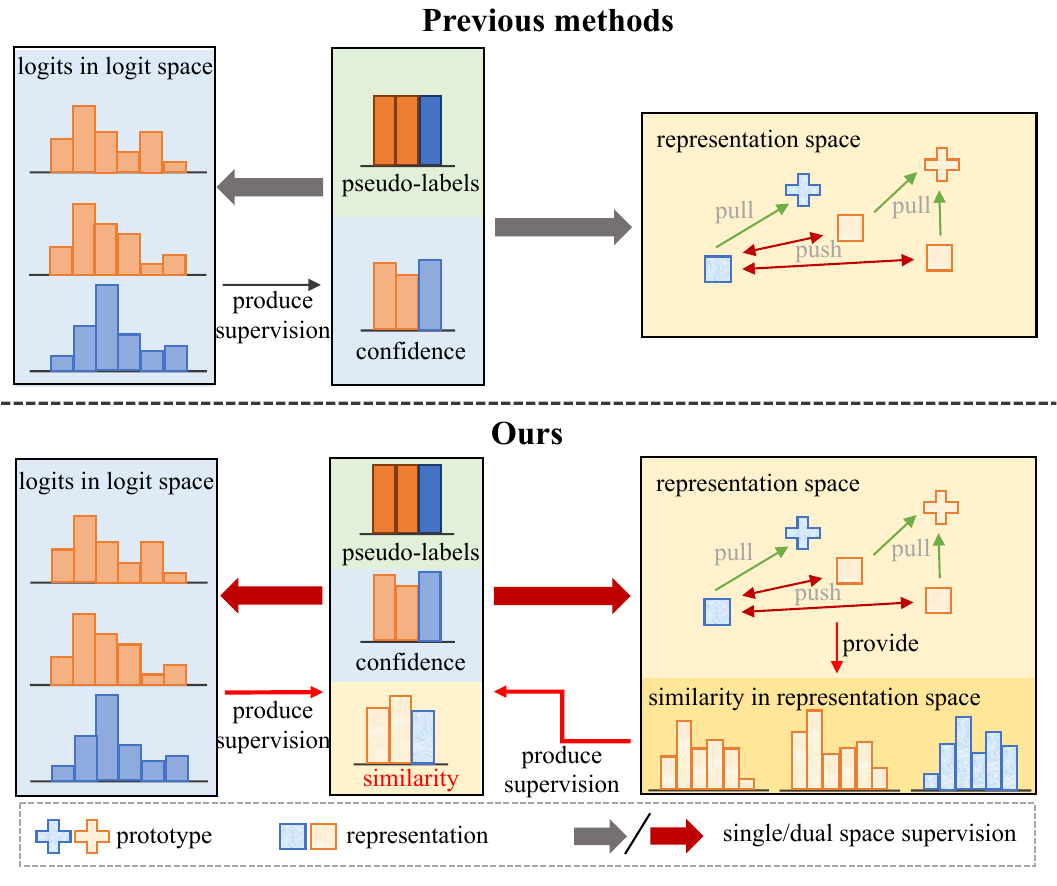}
  \caption{We enhance the knowledge exchange between the logit and representation spaces. Orange and blue represent different classes. Top: Existing contrastive-based S4 methods overlook the semantic information in representation space. Bottom: Our method uses dual-space collaborative supervision.
  }
  \label{cover_fig}
  \vspace{-4mm}
\end{figure}

Most existing works learn from unlabeled images via self-training \cite{MT,classmix,Cutmix} or consistency regularization \cite{cotraining,CCT,UCC} strategies, both of which retrain the model with its predictions on unlabeled images.
Recently, great success has been achieved by introducing pixel-wise contrastive learning to semantic segmentation, which endows the model with a stronger features-extracting ability by accessing a more discriminative representation space.
Specially, these methods \cite{Cipc,RegionContrast} project each pixel to representation space as a representation and regularize it in a fully supervised manner, \textit{i.e.}, aggregating the representations with the same class and separating them with different classes.
In semi-supervised settings, due to limited labels, most methods \cite{CLCMB,Reco,U2PL} obtain supervision from the model's output logits in logit space during the unlabeled training process.
However, recent contrastive-based semantic segmentation methods \cite{CLCMB,Reco,U2PL,Cipc} mainly focus on the learning process in logit space while only taking that in representation space as an auxiliary task.
The unidirectional supervision makes training dominated by the predicted logits, leading to the neglect of information in the representation space. 
We argue that this kind of single-space supervision may incorrectly provide semantic guidance to representation learning and fails to facilitate knowledge exchange between the two spaces (see Sec.~\ref{mix_cross}).

In this work, we extend the single-space supervision to a dual-space supervision for contrastive-based S4 and propose Collaborative Space Supervision (CSS).
Our key insight is to: \textbf{i)} utilize the semantic information in representations to obtain more reliable guidance during unlabeled training, and to enhance knowledge exchange between two spaces; \textbf{ii)} provide a  more accurate reference for the model's performance on predicting each representation to tilt training those under-performing representations.
To achieve objective \textbf{i)}, we obtain dense semantic predictions by retrieving the nearest class prototype for each representation in the representation space and then engage with predictions from the logits for collaborative supervision of the model.
For objective \textbf{ii)}, we measure the similarity between the representations and prototypes and use the similarity after a normalization operation as the indicator for guiding the learning process in the representation space.
Unlike previous works that utilize confidence as the indicator to involve representation learning, the similarity directly reflects the confusion level between representations and prototypes, resulting in more efficient representation learning.

To summarize, our main contributions are three-fold:
\textbf{1)} The dual-space collaboration for contrastive-based S4, enhances the knowledge exchange between the logit and representation spaces.
\textbf{2)} Utilizing similarity to provide a more accurate reference for the model's performance in representation learning.
\textbf{3)} Extensive experiments on two S4 benchmarks demonstrate the effectiveness of our method.

\section{Related Works}
\subsection{Semi-supervised Semantic Segmentation}
The aim of S4 is to train a segmentation model with the semi-supervised setting (\textit{i.e.}, a few labeled images and a large number of unlabeled images) to classify each pixel in an entire image.
The critical issue of S4 is how to leverage unlabeled images to train the model.
Some methods \cite{adversarial4s4,universal_s4,ss_with_gm,s4_with_consistency} based on GANs \cite{gan}, adversarial training \cite{VAT}, and consistency regularization paradigm \cite{CCT,UCC,cotraining,CPS,tcc}.
Meanwhile, self-training \cite{3stage_selftraining,MT,noisy_student,C3-SemiSeg,simple_baseline_for_s4} is also a striking paradigm, which always generates pseudo-labels from model and retrains the model with the combined supervision of human annotations and pseudo-labels.
One essential issue of self-training is the accuracy of pseudo-labels.
Some methods \cite{eln,PSMT,DMT,fixmatch,ST++} try to polish pseudo-labels and provide reliable guidance.
Some methods \cite{DARS,CReST,AEL,subclass_regular} focus on the class-imbalance problems in the dataset and try to alleviate the negative effect from class-biased pseudo-labels generated by the model pre-trained on imbalanced labeled images.
We build our framework based on the self-training and additionally explore semantic information among different images.

\subsection{Pixel-wise Contrastive Learning}
Pixel-wise contrastive learning explores semantic relations not only in the individual image but also among different images.
Different from instance-wise contrastive learning \cite{MoCo,SimCLR,SwAV}, pixel-wise contrastive learning \cite{PixPro,contrast_for_label_efficient,pipa,PC2Seg} project each pixel to the representation in representation space with the cooperation of encoder and representation head.
Representations are then aggregated in their prototypes and are separated from each other in different classes.
In semi-supervised settings, most methods \cite{CLCMB,Reco,U2PL,PRCL} use pseudo-labels based on logits to provide semantic information contrastive learning process during training on unlabeled images.
Meanwhile, the confidence of logit is used as an indicator to involve the contrastive learning process, \eg, \cite{Reco} uses the hard representations whose corresponding logit confidence is lower than a threshold to contrast for effective training.
As opposed to the above methods, we use collaborative space supervision for contrastive learning on unlabeled images and use a new indicator to involve the contrastive learning progress.

\subsection{Prototype-based Learning}
Prototype-based learning has been widely studied in few-shot learning \cite{proto_network,proto_s2,proto_rect_fewshot,dual_prototype_fewshot} and unsupervised domain adaption \cite{proto_uda,proto_contrast_da_s2,proto_framework_uda,trans_uda,Zheng_1}.
Recently, it is restudied in semantic segmentation as known as a non-parametric prototype-based classifier \cite{Rethinking_semantic_segmentation}.
Concretely, the classes in the dataset are presented by a set of non-learnable prototypes, and the dense semantic predictions are thus achieved by assigning the output features to its most similar prototype.
Under semi-supervised settings, some methods \cite{proto_consistency} maintain the consistency between predictions from a linear predictor and a prototype-based predictor.
The two predictors are followed by the encoder and project the features to logit space and representation space, respectively.
In this work, we combine the semantic information in the logit and representation spaces to provide supervision in a collaborative way during semi-supervised learning.

\section{Methodology}
In the S4 task, we have a small labeled set $\mathcal{D}_l=\{(\bm{x}_i^l,\bm{y}_i^l)\}_{i=1}^{N_l}$ and a large unlabeled set $\mathcal{D}_u=\{\bm{x}_i^u\}_{i=1}^{N_u}$, where $\bm{x}_i^l, \bm{x}_i^u$ $\in$ $\mathbb{R}^{H\times W\times 3}$, $H$, $W$ denote the height, and the width, respectively.
And ground truth $\bm{y}_i^l \in \{0,1\}^{H\times W\times \lvert C\rvert}$ with the set of class $C$.
The goal is to boost model performance with $\mathcal{D}_u$.
The base model consists of an encoder $f(\cdot)$ and a segmentation head $g(\cdot)$, which projects features to the logit space $\mathbb{R}^{H\times W\times \lvert C\rvert}$.
We adopt Self-Training (ST) and pixel-wise contrastive learning to our framework, as described in Sec.~\ref{st}.
The supervision for $\mathcal{D}_u$ is produced by the collaboration between the logit and representation spaces, as described in Sec.~\ref{collaborative}.

\subsection{ST and Pixel-wise Contrastive Learning}\label{st}
The main idea of self-training is to pre-train a model on labeled images and use it to produce pseudo-labels as supervision for unlabeled images.
A typical framework is the teacher-student framework \cite{MT}, which consists of a student model and a teacher model.
Both the student model and the teacher model are constructed by an encoder and a segmentation head.
Parameters of the student model are optimized via Stochastic Gradient Decent (SGD) while parameters of the teacher model are updated by the Exponential Moving Average (EMA) of student model parameters.
We denote the encoder and the segmentation head in the student model by $f(\cdot)$ and $g(\cdot)$ while denoting those in the teacher model by $f'(\cdot)$ and $g'(\cdot)$.
The pseudo-labels $\hat{\bm{y}}_i^{u,lgt}$ are produced based on the teacher model's output logits $\hat{\bm{p}}^u_i = g'(f'(\bm{x}^u_i))$ in logit space, formulated as:
\begin{equation}
    \scalebox{0.9}{$ \displaystyle
    \begin{aligned}
        \hat{\bm{y}}_i^{u,lgt} = \bm{1}_c(\mathop{\arg\max}\limits_{c}\{\hat{\bm{p}}_{i,c}^u\}_{c\in C}),
    \end{aligned}
    $}
\end{equation}
where $\bm{1}_c(\cdot)$ denotes the one-hot encoding of class $c$.

In order to enhance the ability of the model itself to extract features, recent works \cite{Reco,U2PL,CLCMB} additionally employ pixel-wise contrastive learning and introduce a representation head to both teacher and student models.
We denote the representation head in the student model as $h(\cdot)$ and that in the teacher model as $h'(\cdot)$.
The pixel $\bm{x}_i$ of the class $c$ are projected as representations $\bm{z}_{ci}$ in representation space by the cooperating of $f(\cdot)$ and $h(\cdot)$, \textit{i.e.}, $\bm{z}_{ci}=h(f(\bm{x}_i))$.
And the representation $\bm{z}_{ci}$ is then aggregated to its class centroid (prototype) while separated from representations in different classes $\bm{z}_{\Tilde{c}i}$ (negatives).
The semantic guidance for contrastive learning is from the combination of ground truth $\bm{y}_i^l$ and pseudo-labels $\hat{\bm{y}}_i^{u,lgt}$ in logit space.
Moreover, in order to emphasize the reliable and crucial aspects during unlabeled and contrastive learning, a sampling strategy is adopted to select valid pixels $\bm{x}_i$ according to their corresponding confidence, \textit{i.e.} the student model's output logits $\bm{p}_i$ after a Softmax operation. 

\noindent\textbf{Discussion.} In recent works \cite{U2PL,CLCMB,Reco}, the supervision of unlabeled images is derived solely from the logit space.
This overlooks the potential benefits of the supervision from the representation space, leading to \textit{two potential limitations}: \textbf{1)} the pseudo-labels $\hat{\bm{y}}_i^{u,lgt}$ obtained from the logit space may contain noise and miss the opportunity to be corrected by semantic information from the representation space; \textbf{2)} since the confidence from logit space is used as the indicator $\hat{j}_i$ for the sampling strategy, learning in the representation space may not be critical enough due to the different confusing parts between the two spaces.

To mitigate these limitations, we produce pseudo-labels from the representation space and combine them with pseudo-labels from the logit space to provide higher-quality supervision during unlabeled training.
Meanwhile, we obtain a new indicator from the representation space for the more effective sampling strategy.
\begin{figure*}[t] 
  \centering
  \includegraphics[width=0.9\linewidth]{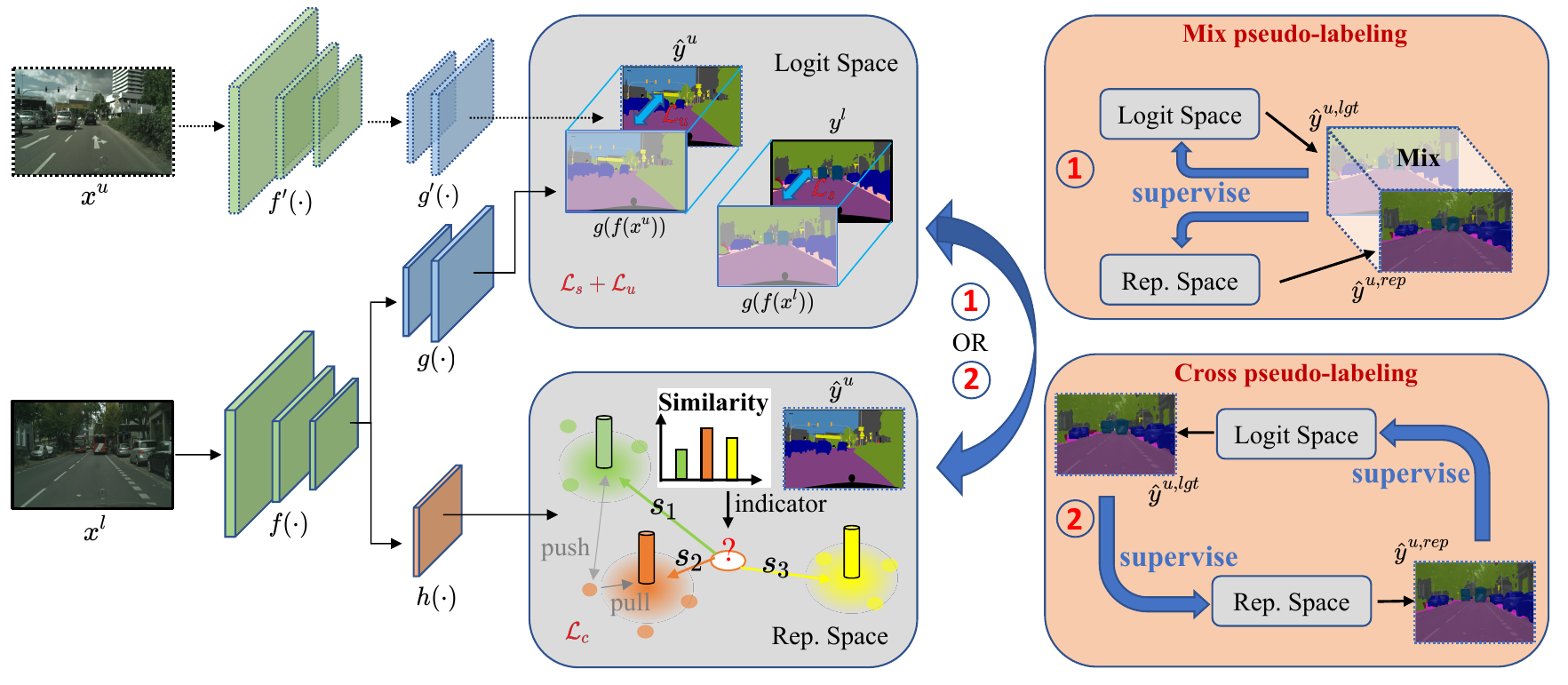}
  \caption{Overview of our framework. Our training pipeline consists of learning in two spaces: logit space and representation space. The pseudo-labels $\hat{\bm{y}}_i^{u}$ during unlabeled training are produced by the collaboration of two spaces with mix pseudo-labeling strategy (1) or cross pseudo-labeling strategy (2). The indicator for representation learning is produced by similarity ($s_1$, $s_2$, and $s_3$).
  }
  \label{framework}
\end{figure*}
\subsection{Supervision from Representation Space}\label{guidance}
In this section, we detail the approach to obtain the pseudo-labels from the representation space.
Meanwhile, we simultaneously access a new indicator for the sampling strategy in representation spaces, which provides a critical reference in the contrastive learning process.

Specifically, we first build a set of prototypes for each class and obtain the pseudo-labels by retrieving the nearest prototype for each representation.
We calculate the centroid of all representations in the current class $c$ as the prototype $\bm{\rho}_c$, which is formulated as:
\begin{equation}\label{eq2}
    \scalebox{0.9}{$ \displaystyle
    \begin{aligned}
        \bm{\rho}_c=\frac{1}{N_c}\sum_i^{N_c}\bm{z}'_i,
    \end{aligned}
    $}
\end{equation}
where $N_c$ is the total number of representations of current class $c$ and $\bm{z}'_i$ is the representation projected by the cooperation of the $f'(\cdot)$ and $h'(\cdot)$.
Meanwhile, to include more representation information, we update the prototype along the sequential iterations with EMA as follows:
\begin{equation}
    \scalebox{0.9}{$ \displaystyle
    \begin{aligned}
        \hat{\bm{\rho}}_c(t)=\alpha \hat{\bm{\rho}}_c(t-1)+(1-\alpha) \bm{\rho}_c(t),
    \end{aligned}
    $}
\end{equation}
where $\hat{\bm{\rho}}_c(t)$, $\hat{\bm{\rho}}_c(t-1)$ mean the current $t^{th}$ prototype and last $(t-1)^{th}$ prototype in iterations , $\bm{\rho}_c(t)$ means the prototype calculated by Eq.~\ref{eq2} in current iteration, and $\alpha$ is a hyper-parameter that controls the updating speed.
Thus, the pseudo-label from the representation space is achieved by:
\begin{equation}
    \scalebox{0.9}{$ \displaystyle
    \begin{aligned}
        \hat{\bm{y}}^{u,rep}_i=\bm{1}_{\hat{c}}(\hat{c}), with~\hat{c}=\mathop{\arg\max}\limits_{c}\{sim(\bm{z}'_i, \bm{\hat{\rho}}_c(t))\}_{c\in C},
    \end{aligned}
    $}
\end{equation}
where $sim(\cdot)$ is defined as the cosine similarity.

As for the indicator for the sampling strategy in the representation space, we use the Softmax function on the similarity among the representation and all prototypes, which is followed as:
\begin{equation}\label{eq9}
    \scalebox{0.9}{$ \displaystyle
    \begin{aligned}
        \hat{j}^{u,rep}_i = \frac{e^{sim(\bm{z_{ci}},\hat{\bm{\rho}}_c}(t))/\tau}{e^{sim(\bm{z_{ci}},\hat{\bm{\rho}}_c(t))/\tau} + \sum_{\Tilde{c}\in \Tilde{C}}e^{sim(\bm{z_{ci}},\hat{\bm{\rho}}_{\Tilde{c}}(t))/\tau}},
    \end{aligned}
    $}
\end{equation}
where $\hat{\bm{\rho}}_{\Tilde{c}}(t)$ means the prototype with different classes from $\bm{z}_{ci}$ and $\tau$ is a hyper-parameter.
Different from using confidence from logit space as the indicator to involve representation learning \cite{Reco,U2PL}, the Softmax similarity directly helps the model to discover the confusion between representations and their prototypes and focus on them during the subsequent training.
\subsection{Collaboration Between Two Spaces}\label{collaborative}
With the pseudo-labels in two spaces, we propose two pseudo-labeling strategies to strengthen the collaboration between two spaces and obtain more reliable pseudo-labels.
\begin{itemize}
    \item \textbf{Mix pseudo-labeling.} To mitigate the negative effects of inherent noise from both two spaces during pseudo-labeling, we adopt the mix pseudo-labeling strategy that only considers the mutually agreeable pseudo-labels between the two spaces.
    Specifically, we define the set of final pseudo-labels as $\hat{Y}^{u}=\hat{Y}^{u,lgt}\cap \hat{Y}^{u,rep}$, where $\hat{\bm{y}}_i^{u,lgt} \in \hat{Y}^{u,lgt}$ and $\hat{\bm{y}}_i^{u,rep} \in \hat{Y}^{u,rep}$.
    \item \textbf{Cross pseudo-labeling.} Inspired by recent researches \cite{CPS,CCT} that maintain consistency among the predictions of the same image across different models or decoders in different views, we propose a cross pseudo-labeling strategy that leverages pseudo-labels from one space to supervise the other. Specifically, we use pseudo-labels $\hat{\bm{y}}_i^{u,rep}$ to supervise the logit space, and vice versa.
\end{itemize}
The strengths of using pseudo-labels from two spaces in the collaborative way are twofold: \textbf{1)} obtaining more reliable supervision during unlabeled training, and \textbf{2)} enabling the strengths of learning in different spaces to complement each other.
Since the learning in different feature spaces concentrates on different parts of features, \textit{i.e.}, the logit space mainly focuses on the most discriminative part of features while the representation space treats all parts of features equally, the performance of pseudo-labels from two spaces varies across different classes or regions of images.
Therefore, our collaborative pseudo-labeling strategies exchange knowledge between two spaces and provide higher-quality supervision during unlabeled training.
The experimental proof is in Sec.~\ref{mix_cross}.

As for indicators, we use confidence as the indicator $\hat{j}_i^{u,lgt}$ for learning logit space and Softmax similarity as the indicator $\hat{j}_i^{u,rep}$ for learning representation space.
We argue that the confusing parts of learning in both two spaces are varied due to the different parts of features being concentrated in each space.
Therefore, this strategy allows the learning in different spaces to focus on their own confusing parts, which can be more effective than using a single indicator when mining confusing parts for both spaces in the training process.
The experimental proof is in Sec.~\ref{ind}.
\subsection{Training Objective}\label{training_obj}
With the indicators $\hat{j}_i^{u,lgt}$ and $\hat{j}_i^{u,rep}$, we adopt some threshold sampling strategies.
In logit space, we set a threshold $\delta_u$ during unlabeled learning and logits $\hat{\bm{p}}_i^u$ whose indicator $\hat{j}_i^{u,lgt}$ is higher than $\delta_u$ will be regarded as the valid logits in logit space.
In representation space, our sample strategy can be divided into three parts: \textbf{1)} \textit{Valid Sampling Strategy}. Similar to the sampling strategy in logit space, a threshold $\delta_w$ is used to sample representations whose indicator $\hat{j}_i^{u,rep}$ is higher than $\delta_w$.
\textbf{2)} \textit{Hard Sampling Strategy}. We adopt the hard sampling strategy for tilting to train those confusing representations.
Specifically, we set a threshold $\delta_s$ to sample representations whose indicator $\hat{j}_i^{u,rep}$ is lower than $\delta_s$.
\textbf{3)} \textit{Negative Sampling Strategy}. We sample negatives according to the similarity between the prototype of current class $c$ and other prototypes.
Concretely, the negatives are more likely to be sampled if its prototype is more similar to the prototype of the current class. 

Cooperated with the ground truth $\bm{y}^l_i$, pseudo-labels $\bm{y}^u_i$ produced from two spaces in a collaborative way, and different sampling strategies in two spaces, the total learning object is composed with a supervised loss $\mathcal{L}_s$, an unsupervised loss $\mathcal{L}_u$, and a contrastive loss $\mathcal{L}_{c}$ as follows:
\begin{equation}
    \scalebox{0.9}{$ \displaystyle
            \mathcal{L} = \mathcal{L}_{s} + \mathcal{L}_{u} + \lambda_{c} \mathcal{L}_{c},
    $}
\end{equation}
where $\lambda_{c}$ is used to tune the contribution between logit space and representation space.
Specifically, $\mathcal{L}_s$ and $\mathcal{L}_u$ are constructed by the Cross-Entropy (CE) $\ell_{ce}$ and can be formulated as:
\begin{equation}
    \scalebox{0.9}{$ \displaystyle
    \mathcal{L}_s = \frac{1}{\lvert \mathcal{B}_l\rvert} \sum_{\bm{x}_i^l\in \mathcal{B}_l} \ell_{ce}(\bm{p}_i^l, \bm{y}^{l}_i),
    $}
\end{equation}
\begin{equation}
    \scalebox{0.9}{$ \displaystyle
    \mathcal{L}_u = \frac{1}{\lvert \hat{\mathcal{B}}_u \rvert } \sum_{\bm{x}_i^u\in \hat{\mathcal{B}}_u} \ell_{ce}(\bm{p}_i^u, \hat{\bm{y}}^{u}_i),
    $}
\end{equation}
where $\mathcal{B}_l$ denotes the labeled images in a mini-batch and $\hat{\mathcal{B}}_u$ is the subsets that sampled from unlabeled images in a mini-batch according to the sampling strategy.
Meanwhile, the contrastive loss $\mathcal{L}_{c}$ is formulated as:
\begin{equation}
    \scalebox{0.9}{$ \displaystyle
    \begin{aligned}
        \mathcal{L}_{c}=& -\frac{1}{|C| \times |\hat{\mathcal{Z}_c}|} \sum_{c \in C}\sum_{\bm{z}_{ci} \in \hat{\mathcal{Z}}_c}\\
        &log[\frac{e^{s(\bm{z}_{ci},\hat{\bm{\rho}}_c)/ \tau}}
        {e^{s(\bm{z}_{ci}, \hat{\bm{\rho}}_c(t)))/ \tau}+\sum_{\tilde{c} \in \tilde{C}} \sum_{\bm{z}_{\tilde{c} i} \in \hat{\mathcal{Z}}_{\tilde{c}}}e^{s(\bm{z}_{ci},\bm{z}_{\tilde{c}i})/ \tau}}],
    \end{aligned}
    $}
\end{equation}
where $\hat{\mathcal{Z}_c}$ is the subset sampled from the set of the representations which belong to class $c$ according to the sampling strategy, $\hat{\mathcal{Z}}_{\tilde{c}}$ is the subset sampled from the set of the representations which bot belong to class $c$, $\Tilde{C}$ denotes the subset of $C$ with class $c$ removed, and the supervision information comes from the final pseudo-labels $\hat{\bm{y}}^{u}_i$ after the pseudo-labeling strategies.
The whole framework is shown in Fig.~\ref{framework}.

\section{Experiments}\label{experiments}
\subsection{Setup}
\noindent\textbf{Datasets.} We conduct experiments on PASCAL VOC 2012 dataset \cite{pascal} and Cityscapes dataset \cite{cityscapes} to validate the effectiveness of our proposed method.
The original PASCAL VOC 2012 dataset contains 1,464 labeled images in \texttt{train} set and 1,449 images for validation in \texttt{val} set.
Following \cite{CPS,classmix}, we additionally introduce 9,118 images from SBD \cite{SBD} as training images.
Since the labels in SBD are coarsely annotated, following \cite{U2PL}, we use both \textit{classic} VOC \texttt{train} set (1,464 candidate labeled images) and \textit{blender} VOC \texttt{train} set (10,582 candidate labeled images).
Cityscapes dataset is a dataset for urban scene understanding, which contains 2,975 images in \texttt{train} set and 500 images in \texttt{val} set.

\noindent\textbf{Network structure.} We use Deeplabv3+ \cite{deeplabv3+} with ResNet-101 \cite{resnet} pre-trained on ImageNet \cite{imagenet} as our network structure.
The segmentation and representation head are composed of \texttt{Conv-BN-ReLU-Conv}.

\noindent\textbf{Implementation details.} For training on PASCAL VOC 2012 dataset, we set the learning rate as 0.0064, weight decay as 0.0005, crop size as 512 $\times$ 512, batch size as 16, and a total of 40,000 iterations.
For training on the Cityscapes dataset, we set the learning rate as 0.0038, weight decay as 0.0005, crop size as 768 $\times$ 768, batch size as 8, and a total of 80,000 iterations.
We use the poly scheduling to decay the learning rate during the training process: $lr=lr_{base}\times(1-\frac{epoch}{total\_epoch})^{0.9}$.
We use the mean of Intersection over Union (mIoU) as the metric in evaluation.
We use the sliding window strategy to evaluate the performance of our method on the Cityscapes dataset, following \cite{CPS}.
In addition, when adopting our cross-labeling strategy, due to the requirement of a set of high-quality prototypes when classifying each representation, we first solely use the supervision from logit space for 20 epochs to initialize the prototypes.

\subsection{Comparison with Existing Methods}
In this subsection, we first reproduce three baselines: MT \cite{MT}, CutMix \cite{Cutmix}, and a typical contrastive-based method with only logit space pseudo-labels and indicators (Baseline) on \textit{classic} VOC \texttt{train} set.
Meanwhile, we make the comparison of our method with mix (CSS (mix)) and cross (CSS (crs.)) pseudo-labeling strategy on \textit{blender} VOC \texttt{train} set and Cityscapes \texttt{train} set with following recent SOTA S4 methods: CCT \cite{CCT}, CPS \cite{CPS}, U$^2$PL \cite{U2PL}, ST++ \cite{ST++}, PRCL \cite{PRCL}, PCR \cite{proto_consistency}, and PSMT \cite{PSMT}.
Since the data split will dramatically affect the performance in S4, \textit{i.e.}, choosing labeled images plays an important role in the final results, we conduct experiments with three different data splits and report the mean value and standard deviation (blue color numbers).
Since the mix pseudo-labeling strategy has better performance, we only use CSS (mix) when compared with SOTAs.
Meanwhile, we use ResNet-101 with deep stem blocks as our network structure when compared with SOTAs.
Since there is no uniform data split, we use the data splits in U$^2$PL \cite{U2PL}.
We use the OHEM loss when training Cityscapes, following \cite{CPS}.

\noindent\textbf{Results on PASCAL VOC 2012.}
Tab.~\ref{tab1} shows the comparison with our baselines on \textit{Classic} PASCAL VOC 2012 set.
Our method consistently outperforms baselines with an acceptable standard deviation on all label rates.
Tab.~\ref{tab2} shows the comparison with the SOTAs on PASCAL VOC 2012.
Our method achieves the best results at most label rates but gets a second at the setting of 1,323 labeled images.

\noindent\textbf{Results on Cityscapes.}
Tab.~\ref{tab3} shows the performance of our method on Cityscapes.
Our method outperforms SOTAs at most label rates but gets a second at the setting of 744 labeled images.
Since multiple prototypes are used in PCR \cite{proto_consistency}, our method is more computationally efficient.
\begin{table}[t]
\small
\centering
\caption{Results on \textit{classic} VOC \texttt{train} set with four different label rate. Labeled data splits are from the original VOC \texttt{train} set. All approaches are reproduced with three runs and three different data splits.} 
\setlength{\tabcolsep}{0.8mm}{%
\begin{tabular}{l|cccc}
\hline
\multicolumn{5}{c}{Pascal VOC 2012 (\textit{Classic})}                                        \\ \hline
Method                             & 92          & 183         & 366         & 732             \\ \hline
Sup.                            & $51.57_{\textcolor{blue}{\pm3.58}}$     & $54.69_{\textcolor{blue}{\pm2.44}}$     & $64.86_{\textcolor{blue}{\pm1.04}}$     & $70.77_{\textcolor{blue}{\pm0.76}}$            \\
MT                              & $58.92_{\textcolor{blue}{\pm2.99}}$     & $61.63_{\textcolor{blue}{\pm1.76}}$     & $66.79_{\textcolor{blue}{\pm0.53}}$     & $71.58_{\textcolor{blue}{\pm0.51}}$           \\
CutMix                          & $65.82_{\textcolor{blue}{\pm3.60}}$     & $67.91_{\textcolor{blue}{\pm1.47}}$     & $72.53_{\textcolor{blue}{\pm0.50}}$     & $74.08_{\textcolor{blue}{\pm0.49}}$            \\
Baseline                        & $66.91_{\textcolor{blue}{\pm4.21}}$     & $70.32_{\textcolor{blue}{\pm1.86}}$     & $73.97_{\textcolor{blue}{\pm1.04}}$     & $76.49_{\textcolor{blue}{\pm0.58}}$            \\ \hline
\rowcolor{gray!20}CSS(crs.)     & $\underline{67.03}_{\textcolor{blue}{\pm4.58}}$     & $\underline{71.41}_{\textcolor{blue}{\pm1.67}}$     & $\underline{74.47}_{\textcolor{blue}{\pm1.08}}$     & $\underline{77.08}_{\textcolor{blue}{\pm0.37}}$      \\
\rowcolor{gray!20}CSS(mix)      & $\bm{68.09}_{\textcolor{blue}{\pm4.89}}$& $\bm{71.93}_{\textcolor{blue}{\pm1.88}}$& $\bm{74.91}_{\textcolor{blue}{\pm1.12}}$& $\bm{77.57}_{\textcolor{blue}{\pm0.73}}$      \\
\hline
\end{tabular}
}
\label{tab1}
\end{table}
\begin{table}[t]
\small
\centering
\caption{Results on \textit{blender} VOC \texttt{train} set. All the results from the recent papers \cite{U2PL,ST++, UCC, eln, PSMT, proto_consistency}. Labeled data is from the augmented VOC \texttt{train} set and the data splits are from \cite{U2PL,PSMT}.}
\setlength{\tabcolsep}{3.4mm}{%
\begin{tabular}{l|cccc}
\hline
\multicolumn{5}{c}{Pascal VOC 2012 (\textit{Blender})}                             \\ \hline
Method                          & 662          & 1323         & 2646         & 5291 \\ \hline
CCT \cite{CCT}                  & $71.86$      & $73.68$      & $76.51$      & $77.40$      \\
CPS \cite{CPS}                  & $74.48$      & $76.44$      & $77.68$      & $78.64$      \\
U$^2$PL \cite{U2PL}             & $77.21$      & $79.01$      & $79.30$      & $80.50$      \\
ST++ \cite{ST++}                & $74.70$      & $77.90$      & $77.90$      & -          \\
PRCL \cite{PRCL}                & $76.96$      & $78.16$      & $79.02$      & $79.59$          \\
PCR \cite{proto_consistency}    & $78.60$      & $\bm{80.71}$ & $80.78$      & $80.91$      \\
PSMT \cite{PSMT}                & $75.50$      & $78.20$      & $78.72$      & $79.76$      \\ \hline 
\rowcolor{gray!20}CSS (mix)     & $\bm{78.73}$ & $\underline{79.54}$ & $\bm{80.82}$ & $\bm{81.06}$      \\\hline
\end{tabular} %
}
\label{tab2}
\end{table}
\begin{table}[t]
\small
\centering
\caption{Results on Cityscapes. The model is trained on the Cityscapes \texttt{train} set, which consists of 2,975 samples in total, and tested on Cityscapes \texttt{val} set. And all the results from the recent papers \cite{U2PL,proto_consistency,PSMT}.}
\setlength{\tabcolsep}{3.4mm}{%
\begin{tabular}{l|cccc}
\hline
\multicolumn{5}{c}{Cityscapes}                             \\ \hline
Method                          & 186          & 372          & 744          & 1488 \\ \hline
CCT \cite{CCT}                  & $69.32$      & $74.12$      & $75.99$      & $78.10$      \\
CPS \cite{CPS}                  & $69.78$      & $74.31$      & $74.58$      & $76.82$      \\
U$^2$PL \cite{U2PL}             & $70.30$      & $74.37$      & $76.47$      & $79.05$      \\
PCR \cite{proto_consistency}    & $73.41$      & $76.31$      & $\bm{78.40}$ & $79.11$      \\ 
PSMT \cite{PSMT}                & -            & $76.89$      & $77.60$      & $79.09$      \\ \hline
\rowcolor{gray!20}CSS (mix)     & $\bm{74.02}$ & $\bm{76.93}$ & $\underline{77.94}$      & $\bm{79.62}$      \\\hline
\end{tabular} %
}
\label{tab3}
\end{table}
\section{Ablative Study}\label{ab}
The main contribution of our work lies in \textbf{1)} collaborative pseudo-labeling strategies and \textbf{2)} a new indicator for representation learning.
To further prove the effectiveness of our proposed method, we conduct ablative studies on these two points.
We choose Deeplabv3+ with ResNet-101 pre-trained on ImageNet as our backbone and leverage 92 labeled images and 183 labeled images in PASCAL VOC 2012.
The other settings are the same as those in Sec.~\ref{experiments}.
\begin{table*}[t]
\small
\centering
\caption{The quality of pseudo-labels from different pseudo-labeling strategies. The pseudo-labels are sampled by sampling strategies.}
\setlength{\tabcolsep}{0.55mm}{%
\begin{tabular}{c|ccccccccccc}
\hline
source & back & aero. & bicy. & bird & boat & bott. & bus & car & cat & chair & cow  \\ \hline
lgt.   & $96.78$     & $95.14$      & $77.47$      & $93.38$     & $81.37$     & $87.54$      & $96.76$    & $95.48$    & $94.47$    & $4.09$      & $92.15$   \\
rep.   & $90.71$     & $96.50$      & $61.42$      & $75.75$     & $53.30$     & $54.65$      & $84.46$    & $80.55$    & $91.11$    & $28.03$     & $88.16$   \\
\hline
mix    & $96.66_{\textcolor{red}{\downarrow0.12}}$     & $98.12_{\textcolor{green!70!black!80}{\uparrow2.98}}$      & $82.76_{\textcolor{green!70!black!80}{\uparrow5.29}}$      & $94.47_{\textcolor{green!70!black!80}{\uparrow1.09}}$     & $85.12_{\textcolor{green!70!black!80}{\uparrow3.75}}$     & $89.94_{\textcolor{green!70!black!80}{\uparrow2.40}}$      & $97.03_{\textcolor{green!70!black!80}{\uparrow0.27}}$    & $95.98_{\textcolor{green!70!black!80}{\uparrow0.50}}$    & $94.65_{\textcolor{green!70!black!80}{\uparrow0.18}}$    & $19.01_{\textcolor{green!70!black!80}{\uparrow14.92}}$      & $94.81_{\textcolor{green!70!black!80}{\uparrow2.66}}$  \\ \hline
source & tabel & dog & horse & motor & pers. & plant & sheep & sofa & train & tv & mIoU  \\ \hline
lgt.   & $58.34$      & $94.12$    & $93.01$      & $91.37$      & $93.68$      & $50.33$      & $91.10$      & $18.16$     & $86.65$      & $64.89$   & $78.87$ \\
rep.   & $52.23$      & $86.97$    & $73.22$      & $88.70$      & $91.75$      & $56.13$      & $66.78$      & $35.25$     & $88.51$     & $62.61$   & $71.57$  \\ 
\hline
mix    & $64.35_{\textcolor{green!70!black!80}{\uparrow6.01}}$      & $94.33_{\textcolor{green!70!black!80}{\uparrow0.21}}$    & $93.65_{\textcolor{green!70!black!80}{\uparrow0.64}}$      & $91.50_{\textcolor{green!70!black!80}{\uparrow0.13}}$      & $93.01_{\textcolor{red}{\downarrow0.67}}$      & $55.64_{\textcolor{green!70!black!80}{\uparrow5.31}}$      & $91.56_{\textcolor{green!70!black!80}{\uparrow0.46}}$      & $29.74_{\textcolor{green!70!black!80}{\uparrow11.58}}$     & $93.54_{\textcolor{green!70!black!80}{\uparrow6.89}}$      & $69.48_{\textcolor{green!70!black!80}{\uparrow4.59}}$   & $82.17_{\textcolor{green!70!black!80}{\uparrow3.33}}$ \\ \hline  
\end{tabular} %
}
\label{tab4}
\end{table*}

\subsection{Effectiveness of Collaborative pseudo-labeling}\label{mix_cross}
\noindent\textbf{Quality of pseudo-labels.} To illustrate the superiority of using pseudo-labels from two spaces in a collaborative way as supervision, we conduct experiments to show the quality of pseudo-labels obtained \textbf{1)} from logit space (lgt.), \textbf{2)} from representation space (rep.), and \textbf{3)} from the mix pseudo-labeling strategy (mix).
The pseudo-labels are sampled with corresponding sampling strategies in Sec.~\ref{training_obj}.
Tab.~\ref{tab4} illustrates the IoU of pseudo-labels for each class on PASCAL VOC 2012 with 92 labeled images.
The results clearly indicate that employing pseudo-labels from the representation space, as opposed to relying solely on those from the logit space, enhances the accuracy of the final pseudo-labels in most classes. This improvement is particularly evident in classes that are originally under-performing, such as the IoU improvement of 11.42$\%$ for the \texttt{chair} class and 11.58$\%$ for the \texttt{sofa} class.

\begin{figure}[t] 
  \centering
  \includegraphics[width=1.0\linewidth]{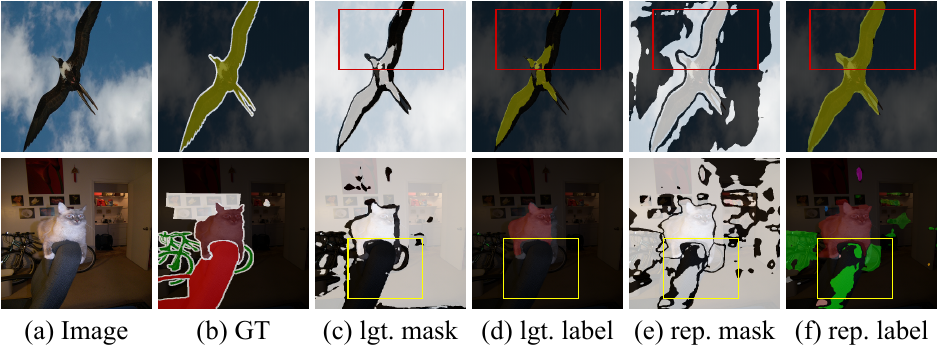}
  \caption{Differences between pseudo-labels in different spaces.
  }
  \label{difference_pseudo_label}
\end{figure}

Meanwhile, we also visualize the pseudo-labels obtained from logit space (lgt.) and representation space (rep.) in Fig.~\ref{difference_pseudo_label}.
Fig.~\ref{difference_pseudo_label} (c) and (e) show the masks for pseudo-labels produced by the sampling strategies.
In particular, the white color represents the valid pixels used during unlabeled learning, while the black color indicates the discarded pixels.
Fig.~\ref{difference_pseudo_label} (d) and (f) are the pseudo-labels we obtained from two spaces.
The figure clearly illustrates the differences between pseudo-labels produced by different spaces.
For example, the parts of the instance edge in pseudo-labels are usually discarded since they are challenging for learning in logit space.
However, pseudo-labels from representation space will easily tackle this problem (shown in the first row).
The pseudo-labels from representation space are more inaccurate in some complex scenes, which be resolved by combining pseudo-labels from logit space (shown in the second row).

We mainly attribute the differences in pseudo-label to the differing concentrations of learning in the two spaces.
Specifically, learning in the logit space primarily emphasizes the most discriminative part of features, whereas that in the representation space treats each part of features equally.
As a result, learning in the logit space may overlook minor feature differences, leading to sub-optimal performance in predicting instance edges and distinguishing between similar classes (\eg, \texttt{chair} and \texttt{sofa}).
Conversely, learning in the representation space produces balanced performance across all image parts and classes.
However, this can lead to erroneous predictions for classes with high intra-class variance (\eg, \texttt{background}).
By leveraging pseudo-labels from the two spaces, we capitalize on the strengths of the learning in each space and enhance the knowledge exchange between the two spaces.

\noindent\textbf{Results of different strategies} To investigate the involvement of different pseudo-labeling strategies, we conduct the experiments as follows: \textbf{1)} Using pseudo-labels from logit space. \textbf{2)} Using pseudo-labels from representation space. \textbf{3)} Using mix pseudo-labeling strategy. \textbf{4)} Using cross pseudo-labeling strategy.
Tab.~\ref{tab5} shows the effectiveness of our proposed strategy in two different label rates.
The results show that the performances of experiments with collaborative pseudo-labeling strategies are better than the ones whose pseudo-labels come from a single space with two different label rates, which proves the effectiveness of our proposed collaboration between the two spaces.
It is worth noting that even though the quality of pseudo-labels from representation space is lower than that from logit space, the performance of the model is also boosted by using the cross pseudo-labeling strategy to maintain consistency between the predictions in two spaces.
In addition, our method with the mix pseudo-labeling strategy outperforms that with the cross pseudo-labeling strategy.
\begin{table}[t]
\small
\centering
\caption{Results on pseudo-labels from different sources on two different label rates.}
\setlength{\tabcolsep}{4.7mm}{%
\begin{tabular}{c|cc}
\hline
\multicolumn{1}{c|}{source}                & 92 labels                                         & 183 labels \\ \hline
logit space                                & $67.11$                                           & $70.32$    \\
representation space                       & $64.20$                                           & $67.52$    \\ \hline
\rowcolor{gray!20}mix pseudo-labeling     & $68.41_{\textcolor{green!70!black!80}{\uparrow1.30}}$               & $72.74_{\textcolor{green!70!black!80}{\uparrow2.42}}$    \\
\rowcolor{gray!20}cross pseudo-labeling   & $67.85_{\textcolor{green!70!black!80}{\uparrow0.74}}$               & $71.98_{\textcolor{green!70!black!80}{\uparrow1.66}}$    \\ \hline
\end{tabular} %
}
\label{tab5}
\end{table}
\begin{figure}[t] 
  \centering
  \includegraphics[width=1.0\linewidth]{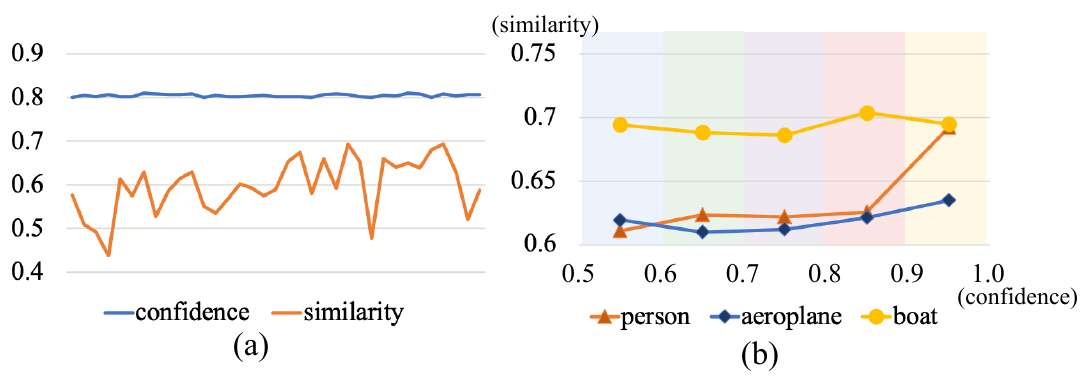}
  \caption{Relations between similarity and confidence.
  }
  \label{conf_and_sim}
  \vspace{-3mm}
\end{figure}
\subsection{Effectiveness of the Indicator}\label{ind}
\noindent\textbf{Limitation of merely using confidence.} To explain the limitation of merely using confidence for learning in the logit and representation spaces, we conduct experiments to show the relations between confidence and similarity.
The similarity is the cosine similarity between representations and prototypes, which directly shows the confusion level between the representation and the prototype of its class.
Fig.~\ref{conf_and_sim} (a) shows the comparison between the confidence of each prediction and the corresponding similarity.
We use the class \texttt{person} for demonstrating in our experiments.
It shows clearly that even though fixing the confidence into a small range (from 0.8 to 0.81 in our settings), the similarity varies.
Meanwhile, in Fig.~\ref{conf_and_sim} (b), different color bars stand for different intervals of confidence, and the lines denote the mean similarity between the prototype and each representation whose corresponding confidence is in the current interval.
Fig.~\ref{conf_and_sim} (b) illustrates that the mean similarity of the class fluctuates when its interval of confidence rises.
Both two figures imply that confidence is not able to represent the confusion level between representations and prototypes since there are no direct and close relations between confidence and similarity.

\begin{figure}[t] 
  \centering
  \includegraphics[width=1.0\linewidth]{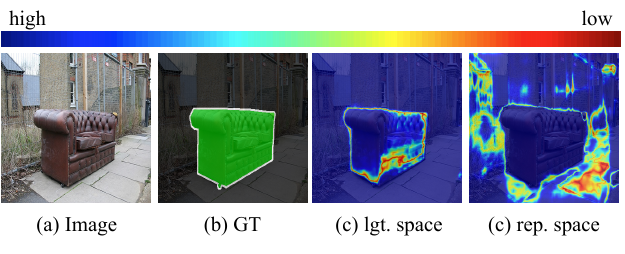}
  \caption{Visualization of the confusing part in different spaces.
  }
  \label{confusing_part}
  \vspace{-4mm}
\end{figure}

Fig.~\ref{confusing_part} visualizes the similarity and confidence of an image in both logit (lgt.) and representation (rep.) spaces, indicating the varying levels of confusion in the same region when learning in different spaces. 

We also attribute it to the different concentrations of learning two spaces, \textit{i.e.}, the confusing region in one space can be more readily addressed in the other space.

Thus, it is inappropriate to choose confidence as the indicator to involve representation learning, \eg, sampling more reliable or hard samples with threshold and indicator for better learning.
In contrast, our indicator directly employs similarity between the representation and prototype of its class, which directly reflects the confusion level in representation learning.
It is more accurate to use similarity as the indicator to sample hard and critical samples in representation learning.

\noindent\textbf{Results of different indicators.} Tab.~\ref{tab6} shows the impact of using different indicators.
We conduct experiments on two different label rates (92 and 183) and three different indicators for pseudo-labels: only confidence (conf.), only similarity (smlr.), and confidence for learning logit space while similarity for learning representation space (mix).
We use two different approaches to obtain pseudo-labels: from logit space only (seg label) and from mix pseudo-labeling strategy (mix label).
It is clear that using both confidence and similarity to involve the learning in their own spaces obtains the best performance, which proves the effectiveness of using different indicators.
\begin{table}[h]
\small
\centering
\caption{Results on indicators from different spaces on two different label rates.}
\setlength{\tabcolsep}{2mm}{%
\begin{tabular}{c|cc|cc}
\hline
           & \multicolumn{2}{c|}{92 labels} & \multicolumn{2}{c}{183 labels} \\ \hline
source                   & seg label                                  & mix label                            & seg label                                  & mix label     \\ \hline
conf.                    & $67.11$                                    & $68.33$                              & $70.32$                                    & $71.92$        \\
smlr.                    & $66.16$                                    & $66.89$                              & $68.90$                                    & $69.25$        \\ \hline
\rowcolor{gray!20}mix    & $67.80_{\textcolor{green!70!black!80}{\uparrow0.69}}$        & $68.41_{\textcolor{green!70!black!80}{\uparrow1.30}}$  & $71.50_{\textcolor{green!70!black!80}{\uparrow1.18}}$        & $72.74_{\textcolor{green!70!black!80}{\uparrow2.42}}$        \\ \hline
\end{tabular} %
}
\vspace{-3mm}
\label{tab6}
\end{table}

\subsection{Ablation study of Components}
In this section, we conduct experiments to introduce our components in CSS step by step, with results shown in Tab.~\ref{tab7}.
Our baseline is the conventional contrastive-based S4, achieving mIoU of $67.11\%$ on 92 labels and $70.32\%$ on 183 labels.
Mix and cross means the pseudo-labels are from the mix pseudo-labeling strategy and cross pseudo-labeling strategy while the indicator is still the confidence in two spaces.
Ind means we use the different indicators in different spaces while the pseudo-labels are from logit space.
The last two rows represent our proposed two pseudo-labeling strategies with indicators from two spaces.
As the result, the mix pseudo-labeling strategy with different indicators boosts model performance by $1.30\%$ and $2.42\%$ while the cross pseudo-labeling strategy with different indicators increases the performance by $0.74\%$ and $1.66\%$.
\begin{table}[ht]
\small
\centering
\vspace{-2mm}
\caption{Ablation study on different components of our CSS.}
\setlength{\tabcolsep}{6.8mm}{%
\begin{tabular}{c|cc}
\hline
component   & 92 labels                           & 183 labels \\ \hline
baseline    & $67.11$                             & $70.32$           \\ \hdashline
mix         & $68.33_{\textcolor{green!70!black!80}{\uparrow1.22}}$ & $71.92_{\textcolor{green!70!black!80}{\uparrow1.60}}$           \\
cross       & $67.21_{\textcolor{green!70!black!80}{\uparrow0.10}}$ & $70.87_{\textcolor{green!70!black!80}{\uparrow0.55}}$           \\
ind         & $67.80_{\textcolor{green!70!black!80}{\uparrow0.69}}$ & $71.50_{\textcolor{green!70!black!80}{\uparrow1.18}}$           \\ \hdashline
mix + ind   & $68.41_{\textcolor{green!70!black!80}{\uparrow1.30}}$ & $72.74_{\textcolor{green!70!black!80}{\uparrow2.42}}$           \\
cross + ind & $67.85_{\textcolor{green!70!black!80}{\uparrow0.74}}$ & $71.98_{\textcolor{green!70!black!80}{\uparrow1.66}}$           \\ \hline
\end{tabular} %
}
\label{tab7}
\vspace{-3mm}
\end{table}
\subsection{Qualitative Results}
\begin{figure}[h] 
  \centering
  \includegraphics[width=1.0\linewidth]{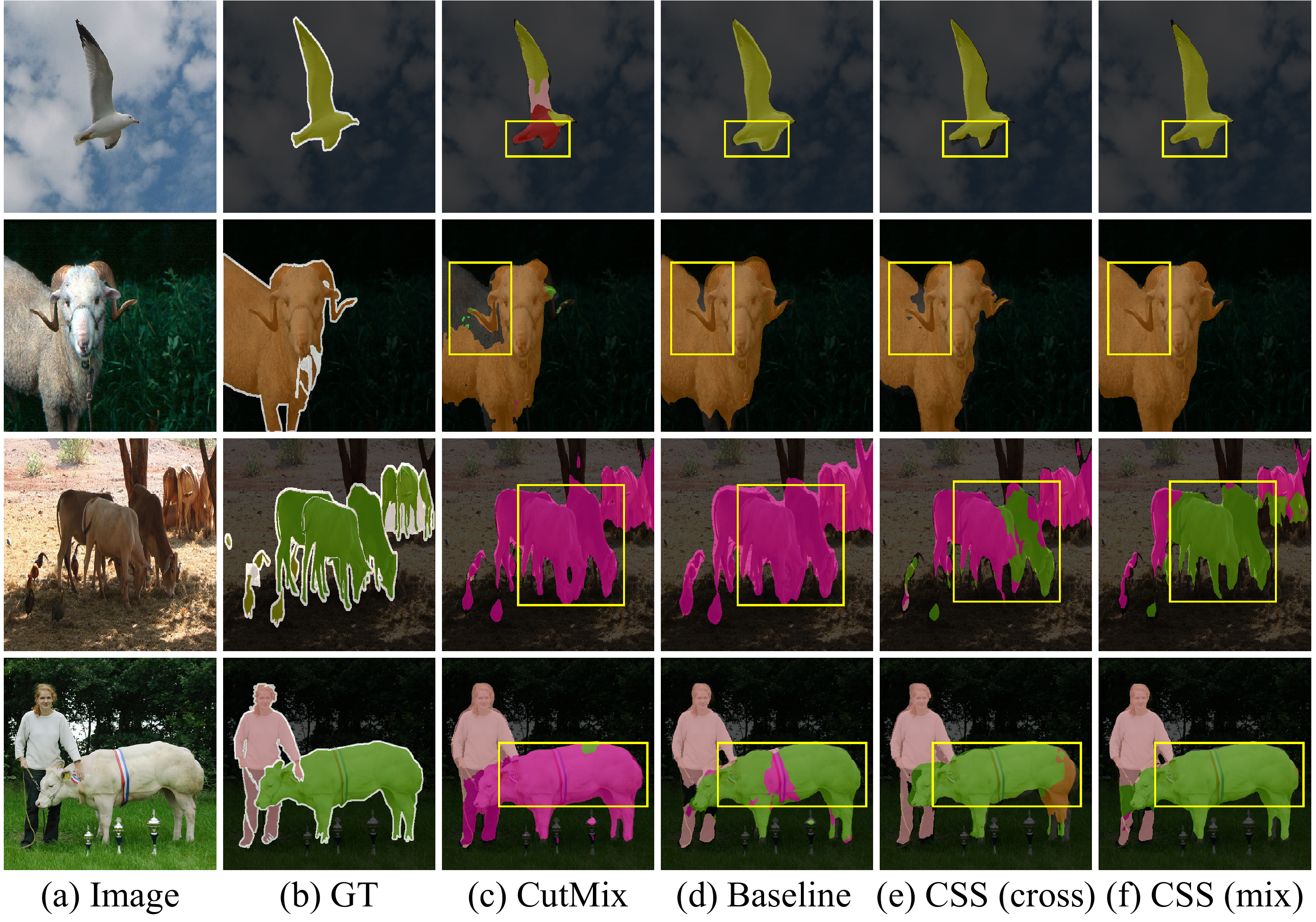}
  \caption{Visualization on PASCAL VOC 2012 with 92 labeled images. Yellow boxes highlight the main differences.
  }
  \vspace{-2mm}
  \label{vis}
\end{figure}

Fig.~\ref{vis} shows the qualitative results of different methods on PASCAL VOC 2012 with 92 labeled images.
Baseline means the conventional contrastive-based method.
Compared with the original self-training methods (CutMix), thanks to introducing pixel-wise contrastive learning, the baseline, and our method perform better in some ambiguous regions.
Furthermore, benefiting from the supervision of two spaces and different indicators in different spaces, our method outperforms the baseline.

\section{Conclusion}
In this paper, we propose two collaborative pseudo-labeling strategies to take full use of the semantic information in the representation space and enhance the knowledge exchange between the logit and representation spaces.
Moreover, we employ a new indicator for the learning process in the representation space.
Extensive experiments demonstrate that our pseudo-labeling strategies obtain more reliable supervision during unlabeled training and our indicator helps the model to concentrate on more critical parts during representation learning.

\noindent\textbf{Future work}: In this paper, we employ pseudo-labeling strategies to utilize the semantic information in both logit and representation spaces.
In the future, we will investigate more powerful strategies to enhance the knowledge exchange between two spaces.

\clearpage
{\small
\bibliographystyle{ieee_fullname}
\bibliography{egbib}
}

\end{document}